\documentclass{article}

\usepackage{arxiv}

\usepackage[utf8]{inputenc}
\usepackage[T1]{fontenc}
\usepackage{hyperref}
\usepackage{url}
\usepackage{booktabs}
\usepackage{amsmath,amssymb,amsfonts}
\usepackage{nicefrac}
\usepackage{microtype}
\usepackage{graphicx}
\usepackage[numbers,sort&compress]{natbib}
\usepackage{doi}
\usepackage{algorithmic}
\usepackage{textcomp}
\usepackage{multirow}
\usepackage{xcolor}
\usepackage{times}
\usepackage{graphicx}
\usepackage{hyperref}

\title{Multimodal Fusion for Fine-Grained Classification of Breast Fibroadenoma and Phyllodes Tumors}

\date{June 2026}

\author{
\small
Chuxi Nan$^{1,2}$, Di Wu$^{2}$, Hongming Guo$^{2}$, Ning Cao$^{1}$,
Xiaohui Zhu$^{1,2,3,*}$, Zhaoting Shi$^{4,5,*}$, and Jiawei Li$^{4,5,*}$\\[0.6ex]
$^{1}$School of Integrated Circuits, Wuxi Vocational College of Science and Technology\\
$^{2}$School of Artificial Intelligence and Advanced Technology, Xi'an Jiaotong-Liverpool University\\
$^{3}$Shanghai Gem Science and Technology Co., Ltd.\\
$^{4}$Department of Oncology, Shanghai Medical College, Fudan University\\
$^{5}$Department of Medical Ultrasound, Fudan University Shanghai Cancer Center\\[0.6ex]
$^{*}$Corresponding authors: Xiaohui Zhu, Zhaoting Shi, and Jiawei Li.
}

\renewcommand{\shorttitle}{}

\hypersetup{
pdftitle={Multimodal Fusion for Fine-Grained Classification of Breast Fibroadenoma and Phyllodes Tumorsxiv style},
pdfsubject={q-bio.NC, q-bio.QM},
pdfauthor={Chuxi Nan},
pdfkeywords={First keyword, Second keyword, More},
}

\begin{document}
\maketitle

\thispagestyle{empty}
\pagestyle{plain}

\footnotetext{This work involved human subjects in its research. Approval of all ethical and experimental procedures and protocols was granted by Fudan University Shanghai Cancer Center, China.}


\begin{abstract}
Fibroadenoma (FA) and phyllodes tumor (PT) are both fibroepithelial lesions of the breast and exhibit highly overlapping features on ultrasound imaging. Consequently, even experienced sonographers achieve only a 20--30\% success rate in differentiating FA, benign PT, and borderline PT. However, because these three entities require different treatment strategies, accurate preoperative differentiation is particularly important. Existing studies have mainly focused on benign--malignant classification, while fine-grained classification tasks and corresponding multimodal data resources remain limited. To address this gap, we constructed FAPT-M, the world's first multimodal breast ultrasound dataset specifically designed for the fine-grained classification of FA, benign PT, and borderline PT. Based on this dataset, we further propose a clinically conditioned multimodal learning framework built upon a condition--interact--retain information flow. The proposed method achieves 78.06\% accuracy, 72.63\% F1-score, and 89.77\% AUC on FAPT-M, outperforming the evaluated representative image-based and multimodal methods overall. Ablation studies further validate the model design. In addition, supplementary experiments on the general-purpose multimodal PetFinder dataset further demonstrate the generalizability of the proposed method. The FAPT-M dataset is now publicly available, and the implementation code has also been released on 
\href{https://github.com/Lilpumpkiin/Multimodal-Fusion-for-Fine-Grained-Classification-of-Breast-Fibroadenoma-and-Phyllodes-Tumors}{GitHub}.

\end{abstract}


\section{Introduction}
\label{sec:introduction}

Breast fibroadenoma (FA) and phyllodes tumor (PT) are fibroepithelial breast tumors with substantially overlapping clinical and imaging characteristics. FA is one of the most common benign breast tumors, whereas PT is relatively rare and can be classified into benign, borderline, and malignant subtypes \cite{Kalambo2018Phyllodes}. FA and PT exhibit highly overlapping ultrasound features, making reliable preoperative differentiation challenging even for experienced sonographers. Clinically, FA is usually managed conservatively or by simple excision \cite{Greenberg1998Management}, whereas PT generally requires wider local excision because of its higher recurrence potential \cite{Adesoye2016Current}. Therefore, accurate preoperative differentiation between FA and non-malignant PT is important for treatment planning and surgical decision-making.

Deep learning has been extensively investigated in breast ultrasound imaging \cite{Alotaibi2023BreastCNN,Dan2024BUSDeepLearningReview}. However, most existing studies focus on relatively coarse-grained tasks, such as benign--malignant classification \cite{Hu2025Performance,Mashekova2025Review}, while fine-grained differentiation among histologically related fibroepithelial tumors remains underexplored. Existing studies often frame FA--PT differentiation as a binary classification problem or merge different PT subtypes into a single category \cite{Stoffel2018Distinction}, overlooking heterogeneity among PT subtypes and subtle imaging differences between lesions. Distinguishing FA, benign PT, and borderline PT is therefore both clinically meaningful and technically challenging \cite{Niu2021Differential}. Moreover, the limited availability of standardised multimodal FA--PT datasets further constrains the development and evaluation of multimodal learning approaches \cite{Lu2025IntraTumor,Abir2025MultimodalDL}.  In real-world clinical diagnosis, clinicians typically make decisions by jointly considering ultrasound images, descriptive reports, and structured clinical information; however, many current deep learning methods rely predominantly on imaging features alone \cite{Chen2025DeepLearning}.

To address these challenges, we construct FAPT-M, a pathology-confirmed multimodal cohort for the fine-grained classification of FA, benign PT, and borderline PT. Each sample integrates a B-mode ultrasound image, an ultrasound diagnostic description, and structured clinical variables. Using this dataset, we investigate a clinically conditioned multimodal learning framework that preserves modality-specific evidence while incorporating patient-specific clinical context. Rather than directly concatenating multimodal representations, we first condition the image and text representations on structured clinical variables, after which the conditioned representations interact with the clinical representation through modality-token interaction. The classifier then retains access to both the original modality representations and their interaction-updated representations, forming a condition--interact--retain information flow.

We evaluate the proposed framework using fixed patient-disjoint five-fold cross-validation with separate training, validation, and test partitions. We compare against representative image-based and multimodal baselines and conduct module- and modality-level ablation studies to assess the contributions of clinical conditioning, multimodal interaction, and raw--contextual representation retention.

The main contributions of this study are summarised as follows:

\begin{itemize}

\item We established FAPT-M, the world’s first multimodal dataset designed to distinguish fibroadenomas, benign phyllodes tumors and borderline phyllodes tumors and made it publicly available to support future research on fibroepithelial breast tumors.

\item We propose a clinically conditioned multimodal learning framework featuring a ‘conditioning–interaction–retention’ information flow. Structured clinical variables modulate image and text representations before multimodal interaction. Subsequently, the conditioned image and text representations interact with the clinical representations, whilst the classifier simultaneously retains both the original representations and the interaction-updated representations for final prediction.

\item In experiments, the proposed framework outperformed representative image-based and multimodal baseline models. Ablation studies at both the module and modality levels further confirmed the contributions of clinical conditioning, multimodal interaction, and retaining both original and context-updated representations.

\end{itemize}

\section{Related Work}

\subsection{Ultrasound-based Breast Tumor Diagnosis}

Breast ultrasound (BUS) is widely used for tumor screening because it is radiation-free, low-cost, and reproducible in dense breast tissue \cite{Iacob2024Evaluating}. However, fibroadenoma (FA) and phyllodes tumor (PT) overlap substantially in B-mode ultrasound features, making differentiation challenging \cite{Wiratkapun2014Fibroadenoma,Duman2016Differentiation,Kalambo2018Phyllodes}. Recent deep learning methods have improved BUS analysis, but most focus on benign--malignant classification or general lesion detection rather than FA--PT discrimination \cite{Wilding2022Deep,Alotaibi2023BreastCNN,Dan2024BUSDeepLearningReview}. Existing FA--PT studies typically simplify the task by merging PT subtypes into a single category \cite{Stoffel2018Distinction,Yan2024DeepLearning,Lu2025IntraTumor}. In contrast, this work focuses on fine-grained classification of FA, benign PT, and borderline PT using multimodal data. Pathology-confirmed multimodal datasets specifically designed for the fine-grained differentiation of FA, benign PT, and borderline PT remain limited.

\subsection{Multimodal Learning in Medical Imaging}

Multimodal learning combines heterogeneous medical information and can provide complementary evidence when lesions have ambiguous imaging appearances \cite{Li2024ReviewFusion,Cui2023DeepFusion}. Existing breast studies have fused multiple imaging views or modalities and combined images with clinical information \cite{Chen2025DeepLearning,Wei2025MultimodalBreast,Muhtadi2025MultimodalBreast}. Among the publicly available methods considered in this study, few directly match our exact combination of ultrasound images, Chinese report text, and structured clinical measurements. LMF provides a general low-rank fusion mechanism \cite{Liu2018LMF}, MMBCD was designed for mammography and clinical-history text \cite{Jain2024MMBCD}, and AutoMM supports image, text, and tabular inputs \cite{Tang2024AutoMM}. Consequently, LMF and MMBCD require task- and input-specific adaptation, while all three require fine-tuning on our dataset. AutoMM-LF denotes our configured late-fusion baseline rather than a separately published model. Our framework differs by conditioning image and text representations on clinical information before interaction and retaining both original and contextualized modality states.

\subsection{Feature Encoding and Conditional Interaction}

CNNs remain widely used in medical image analysis because they effectively capture local texture and structural patterns \cite{Salehi2023CNNReview,Sistaninejhad2023DeepReview}. DenseNet improves feature reuse and gradient propagation through dense connectivity \cite{Huang2017DenseNet}, and has been applied to breast ultrasound, mammography, and histopathological analysis\cite{AlZoubi2024BreastDL,Liao2023MammoAsymmetry,Li2020IDSNet}. In our framework, DenseNet121 serves as the visual encoder, while the pretrained Chinese-CLIP text encoder provides representations of Chinese ultrasound descriptions\cite{Yang2022ChineseCLIP}. FiLM performs feature-wise affine transformation using conditioning information \cite{Perez2018FiLM}, while Transformer self-attention models relationships among modality tokens \cite{Vaswani2017Transformer}. Both are established techniques. Our contribution lies in their task-specific organization: clinical variables condition the image and text representations before interaction, remain available as an explicit clinical token, and support classification using both original and contextualized modality representations.

\section{Dataset Construction and Data Preparation}

\begin{figure}[!t]
\centerline{\includegraphics[width=0.60\columnwidth]{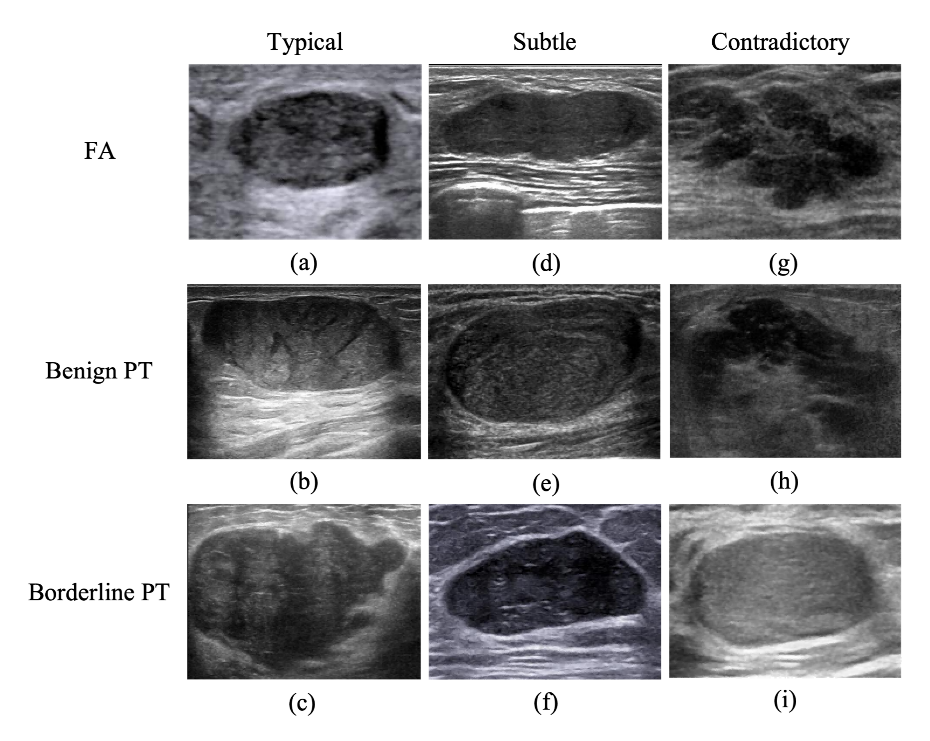}}
\caption{Representative B-mode ultrasound images of FA and PT. (a) FA with regular shape and homogeneous echotexture. (b) Benign PT with relatively circumscribed margin and heterogeneous echotexture. (c) Borderline PT with lobulated margin and heterogeneity. (d)--(f) cases with subtle differences but different pathology. (g)--(i) atypical presentations increasing diagnostic difficulty.}
\label{fig1}
\end{figure}

FA and PT exhibit substantial similarity on ultrasound, although subtle structural differences exist among different tumor subtypes \cite{Kamitani2014,Lohitvisate2024,Duman2016Differentiation}. These overlapping imaging characteristics make preoperative differentiation challenging, particularly for distinguishing benign PT and borderline PT from FA. Representative cases are shown in Fig.~\ref{fig1}. These differences typically change gradually rather than forming clearly separable class boundaries. Therefore, accurate and consistent annotation is particularly important for fine-grained classification.

In clinical practice, breast sonographers do not rely solely on ultrasound images but also consider clinical information such as patient age, tumor size, and lesion growth characteristics \cite{Wu2012}. Such information can provide complementary clues regarding tumor biology and assist in the differentiation of fibroepithelial lesions \cite{Reis2021,Wiratkapun2014Fibroadenoma}. Therefore, using ultrasound images alone as training data may not fully reflect the actual clinical diagnostic process.

\subsection{Dataset Annotation and Data Collection}

All samples in the FAPT-M dataset were verified by pathological examination. Multiple experienced breast sonographers cross-reviewed the annotations to ensure consistency and reliability. Each sample consists of one ultrasound image, a corresponding Chinese ultrasound description, and four structured clinical variables, including patient age and three lesion dimensions.

This retrospective study was approved by Fudan University Shanghai Cancer Center, and the requirement for informed consent was waived because of the retrospective study design. We anonymized all data before analysis.

The final dataset includes 3,600 ultrasound images from 876 patients: 1,786 images from 410 patients with FA, 787 images from 211 patients with benign PT, and 1,027 images from 255 patients with borderline PT. Patient age ranged from 14 to 88 years, with a mean age of $41.17\pm12.07$ years and a median age of 41 years. Postoperative pathological examination served as the reference standard and was not provided as an input to the model.

\begin{table}[!t]
\centering
\caption{Demographics and clinical characteristics of patients.}
\label{tab:patient_demographics}
\renewcommand{\arraystretch}{1.1}
\resizebox{0.50\linewidth}{!}{
\begin{tabular}{lccc}
\toprule
Characteristics & FA & B & BD \\
\midrule
Number & 410 & 211 & 255 \\
Ages(Years) & $40.15\pm12.47$ & $40.13\pm11.05$ & $43.70\pm11.84$ \\
Max diameter(mm) & $18.50\pm8.14$ & $27.84\pm10.68$ & $33.97\pm12.07$ \\
BI-RADS Categories: & & & \\
\quad 3 & 63 & 47 & 49 \\
\quad 4A & 330 & 134 & 164 \\
\quad 4B & 16 & 27 & 37 \\
\quad 4C & 1 & 3 & 5 \\
\bottomrule
\end{tabular}
}

\vspace{2pt}
\scriptsize\raggedright
\footnotesize Note: FA = fibroadenoma; B = benign phyllodes tumor; BD = borderline phyllodes tumor; BI-RADS = Breast Imaging Reporting and Data System. BI-RADS categories indicate standardized breast imaging assessment levels, with higher categories generally reflecting greater suspicion of malignancy or the need for further clinical management \cite{ACR2013BIRADS}.
\end{table}

We excluded malignant PT cases from the present study because only 33 such patients were collected during the study period, resulting in a substantially smaller sample size than the other three categories. Including this limited subgroup would introduce severe class imbalance and make reliable model training and evaluation difficult. In addition, malignant PT may present with relatively more distinct clinical and imaging characteristics than FA, benign PT, and borderline PT. Therefore, the present study focuses on the more challenging fine-grained differentiation among FA, benign PT, and borderline PT.

\subsection{Ultrasound Image Acquisition}

Ultrasound images were acquired using multiple clinical ultrasound systems, including Philips EPIQ7 and iU22, GE LOGIQ E9, V730, and LOGIQ S8, Mindray Resona 7, and Toshiba Aplio 500. Linear-array transducers operating at frequencies of 5--12 MHz were used for image acquisition. For larger lesions, panoramic or trapezoidal imaging modes were applied when necessary to preserve the complete lesion region.

\section{Method}

Given the characteristics of breast fibroadenoma and the clinical diagnostic process that integrates ultrasound images, diagnostic descriptions, and structured clinical information, we propose a clinically conditioned multimodal learning framework. The proposed framework aims to exploit complementary information across modalities while preserving modality-specific diagnostic evidence that supports fine-grained classification.

Unlike conventional multimodal approaches that directly concatenate or fuse representations from different modalities, we design a condition--interact--retain information flow mechanism. First, structured clinical variables are utilized as conditioning signals to adaptively modulate image and text representations before multimodal fusion, enabling modality representations to incorporate patient-specific clinical context. Subsequently, the clinically conditioned image and text representations, together with the clinical representation, are fed into a cross-modal interaction module, where attention mechanisms among modality tokens are employed to capture complementary relationships across heterogeneous information sources. Finally, during classification, the framework retains both the original modality representations and the interaction-updated contextual representations, reducing the potential loss of modality-specific diagnostic evidence during fusion.

\begin{figure*}[!t]  
\centering           
\includegraphics[width=\textwidth]{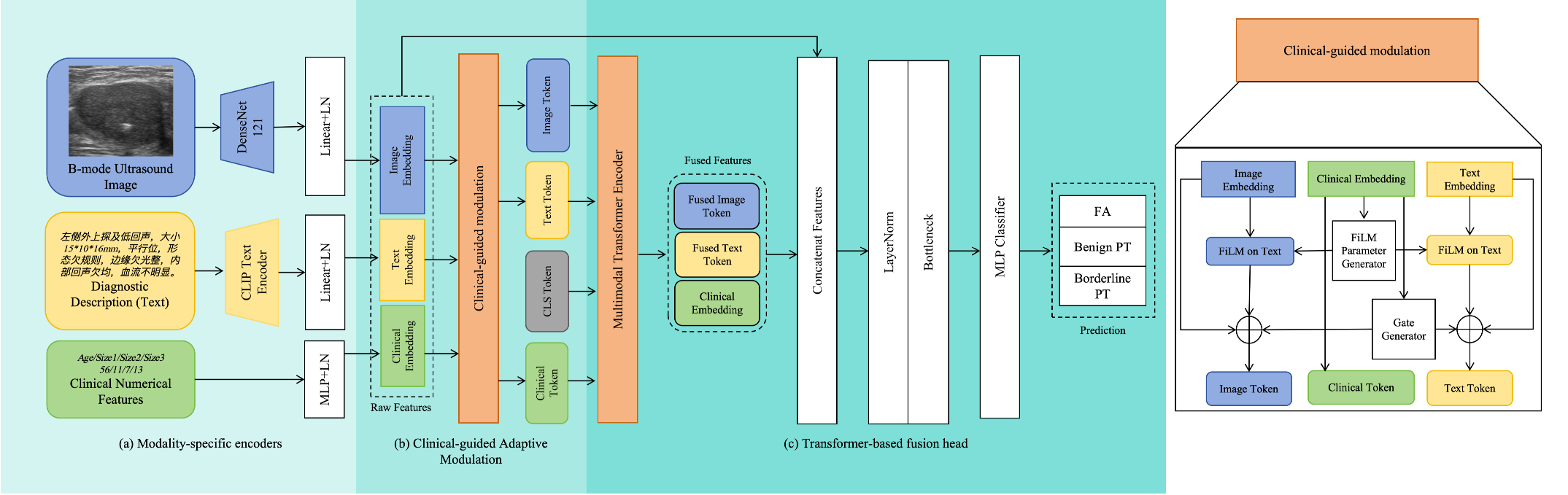}  
\caption{Overview of the proposed framework. (a) Modality-specific encoders extract image, text, and structured clinical representations. (b) The clinical representation generates branch-specific modulation parameters and modality-level gates to condition the image and text representations. (c) A modality-token Transformer models joint interactions, after which both the original and contextualized modality representations are retained for three-class classification.}
\label{fig2}
\end{figure*}

As illustrated in Fig. \ref{fig2}, the proposed framework consists of three main components: (1) modality-specific feature encoding module; (2) clinical-conditioned representation modulation module; and (3) cross-modal interaction and raw--contextual representation retention module.

\subsection{Modality Representation Learning}

Because ultrasound images, diagnostic descriptions, and structured clinical variables differ in nature, we first use modality-specific encoders to extract representations from each modality independently. The extracted representations are projected into a shared embedding space to enable subsequent clinical conditioning and cross-modal interaction.

For each patient sample, three modality representations are obtained:

\[
\mathbf{z}_{\mathrm{img}},
\mathbf{z}_{\mathrm{txt}},
\mathbf{z}_{\mathrm{clin}}
\in
\mathbb{R}^{512},
\]

where $\mathbf{z}_{\mathrm{img}}$, $\mathbf{z}_{\mathrm{txt}}$, and $\mathbf{z}_{\mathrm{clin}}$ denote the ultrasound image representation, diagnostic text representation, and structured clinical representation, respectively.

The image and text modalities are encoded using pretrained modality-specific encoders, while the clinical representation is obtained from structured clinical variables, including patient age and three lesion dimensions. This stage aims to achieve dimensional compatibility among heterogeneous modalities rather than explicitly optimize cross-modal semantic alignment.

The obtained modality representations provide the initial evidence for subsequent clinical-conditioned modulation and cross-modal interaction.

\subsection{Clinical-conditioned Representation Modulation}

In clinical diagnosis, identical ultrasound appearances may correspond to different diagnostic interpretations depending on patient-specific factors, such as age and lesion characteristics. Therefore, rather than treating clinical variables as an additional input at the final classification stage, we use clinical information as a conditioning signal to adjust image and text representations before multimodal interaction adaptively.

Given the clinical representation $\mathbf{z}_{\mathrm{clin}}$, modality-specific modulation parameters are generated independently for image and text branches:

\[
[\boldsymbol{\gamma}_{m},\boldsymbol{\beta}_{m}]
=
W_m\mathbf{z}_{\mathrm{clin}}+\mathbf{b}_m,
\quad
m\in\{\mathrm{img},\mathrm{txt}\},
\]

where $\boldsymbol{\gamma}_{m}$ and $\boldsymbol{\beta}_{m}$ represent feature-wise scaling and shifting parameters, respectively.

The conditioned representation is obtained as:

\[
\mathbf{v}_{m}
=
\boldsymbol{\gamma}_{m}
\odot
\mathbf{z}_{m}
+
\boldsymbol{\beta}_{m},
\]

where $\odot$ denotes element-wise multiplication.

To control the influence strength of clinical information for different modalities, a sample-specific gating mechanism is introduced:

\[
\mathbf{g}_{m}
=
\sigma(W_g\mathbf{z}_{\mathrm{clin}}+\mathbf{b}_g),
\]

where $\sigma(\cdot)$ denotes the sigmoid activation function.

The final clinically conditioned modality representation is formulated as:

\[
\widetilde{\mathbf{z}}_{m}
=
\mathbf{z}_{m}
+
\mathbf{g}_{m}\mathbf{v}_{m}.
\]

The residual connection preserves the original modality evidence, while the gating mechanism dynamically controls the contribution of clinical context.

Through this process, clinical variables are transformed from auxiliary information into active conditioning signals that guide modality-specific representation learning.

\subsection{Cross-modal Interaction and Evidence Retention}

After clinical conditioning, the image, text, and clinical representations are jointly incorporated into a cross-modal interaction module.

The input token sequence is constructed as:

\[
\mathbf{H}^{0}
=
[
\mathbf{z}_{cls},
\widetilde{\mathbf{z}}_{img},
\widetilde{\mathbf{z}}_{txt},
\mathbf{z}_{clin}
]
+\mathbf{P},
\]

where $\mathbf{z}_{cls}$ is a learnable classification token and $\mathbf{P}$ denotes positional embeddings.

A Transformer encoder is then employed to model relationships among different modality tokens:

\[
\mathbf{H}^{L}
=
Transformer(\mathbf{H}^{0}).
\]

The interaction-updated contextual representations are denoted as:

\[
\mathbf{h}_{img},
\mathbf{h}_{txt},
\mathbf{h}_{clin}.
\]

Different from approaches that only use clinical information as an auxiliary feature, our clinical representation plays two complementary roles:

(1) it acts as a conditioning signal to modulate image and text representations before interaction;

(2) it serves as an explicit clinical token during cross-modal interaction.

To avoid excessive information compression and potential loss of modality-specific evidence during fusion, both original and contextualized representations are retained:

\[
\mathbf{r}
=
[
\mathbf{z}_{img},
\mathbf{z}_{txt},
\mathbf{z}_{clin},
\mathbf{h}_{img},
\mathbf{h}_{txt},
\mathbf{h}_{clin}
].
\]

The final representation $\mathbf{r}$ is subsequently fed into a bottleneck layer and classifier for three-class prediction.

By jointly utilizing original modality evidence and cross-modal contextual information, the proposed framework establishes a complete
\textbf{condition--interact--retain} multimodal information flow.

\section{Experiments}

\subsection{Implementation Details}

All experiments in this research, including comparative and ablation experiments, employed a fixed, patient-level, mutually exclusive five-fold cross-validation strategy for model training and evaluation. In each experiment, three folds of the five-fold data were used for training, one for validation, and one for testing, ensuring that each fold served as the test set exactly once. All comparison methods used the same predefined training, validation, and test splits to avoid unfair comparisons due to differences in data partitioning. All continuous clinical variables were z-score standardised using only the mean and standard deviation of the corresponding training set, and the same standardisation parameters were applied to both the validation and test sets to prevent data leakage.

All models were trained for 100 epochs, with an initial learning rate of $1\times10^{-4}$, a batch size of 32, and optimisation using the cross-entropy loss function. The experiments ran on an NVIDIA RTX 4090 GPU (24 GB), using Python 3.12, PyTorch 2.5.1, CUDA 12.4, and Ubuntu 22.04. The computing platform was equipped with a 16-core Intel Xeon Platinum 8358P CPU and 120 GB of RAM.

\subsection{Evaluation Metrics}

To comprehensively evaluate the model’s performance in a three-class classification task, this paper employs accuracy (ACC), macro-precision (PRE), macro-recall (REC), macro-F1-score, macro-sensitivity, macro-specificity, macro-average area under the ROC curve (AUC) and negative log-likelihood (NLL) as evaluation metrics.

For PRE, REC and F1-score, the metrics for each category are first calculated individually, then macro-averaged after assigning equal weights to each category. Sensitivity and Specificity are calculated separately using the one-vs-rest approach and then averaged across categories. Under the current definition, macro-averaged Sensitivity equals macro-averaged Recall; to maintain consistency with the experimental tables, this paper reports both metrics.

The AUC is calculated using the one-vs-rest approach based on the model’s predicted probabilities and is then macro-averaged. ACC represents the proportion of correctly classified samples out of the total. NLL denotes the average negative logarithm of the probabilities assigned by the model to the true classes; a lower value indicates better probability prediction performance. Except for NLL, all other metrics are reported as percentages. All compared methods adopt the same definitions for these metrics.

\subsection{Comparative Experiments}
\label{sec:comparative_experiments}

To evaluate the proposed method’s fine-grained classification capabilities on FAPT-M, we compared it with representative image classification models and independent multimodal learning methods.

\subsubsection{Comparison on FAPT-M Dataset}

First, we compare the proposed method with a range of representative image classification models, including Swin Transformer \cite{Liu2021SwinTransformer}, ResNet, DenseNet \cite{Huang2017DenseNet}, EfficientNetV2 \cite{Tan2021EfficientNetV2}, ViT\cite{dosovitskiy2021vit}, DeiT \cite{touvron2021deit} and CLIP\cite{radford2021clip}. This experiment is designed to evaluate the performance differences between models that rely solely on ultrasound images and our proposed comprehensive multimodal framework.

The results of the comparison with image-based baselines are shown in Table~\ref{tab:image_baselines}. The proposed method achieved an ACC of 78.06\%, an F1-score of 72.63\% and an AUC of 89.77\%, demonstrating the best overall classification performance among the evaluated methods. Compared with the better-performing image-based baselines, the comprehensive multimodal framework demonstrated advantages across all key evaluation metrics.

In addition to the image baseline, we selected three independent multimodal methods for comparison: LMF \cite{Liu2018LMF}, MMBCD \cite{Jain2024MMBCD}, and AutoMM-LF \cite{Tang2024AutoMM}, a late-stage fusion model built on AutoMM.

LMF represents a classical low-rank multimodal fusion method \cite{Liu2018LMF}; MMBCD is a medical multimodal model designed for breast imaging tasks, with mammography images and clinical history text as its original inputs \cite{Jain2024MMBCD}; AutoMM-LF represents a general-purpose multimodal learning framework supporting images, text and structured features \cite{Tang2024AutoMM}. Because the original tasks and input formats of these methods do not fully match the multimodal format in this study, we adapted the inputs to suit the current task; in particular, AutoMM-LF refers to the advanced fusion configuration built on AutoMM in this paper, rather than a separately published method.

The experimental results for the independent multimodal baselines are shown in Table~\ref{tab:multimodal_baselines}. Among the three baselines, LMF achieved the highest ACC, F1-score and AUC, at 70.82\%, 62.60\% and 81.18\%, respectively. In comparison, our method achieved 77.64\%, 73.38\% and 89.74\%, respectively, representing improvements of 6.82\%, 10.78\% and 8.56\%points.

It should be noted that whilst our method achieves higher results on the primary classification metrics, NLL is not optimal in all comparisons. For example, LMF’s NLL is lower than our method's. Therefore, we report classification performance and probability prediction performance separately, rather than using a single metric to conclude overall performance.

\begin{table*}[!t]
\centering
\caption{Comparative Experiment Results of image-based baselines on the FAPT-M Dataset}
\label{tab:image_baselines}
\fontsize{8}{10}\selectfont
\setlength{\tabcolsep}{4pt}
\begin{minipage}{\linewidth}
\centering
\begin{tabular}{l c c c c c c c c}
\toprule
Method & ACC(\%) & F1-Score(\%) & PRE(\%) & REC(\%) & Sensitivity(\%) & Specificity(\%) & AUC(\%) & NLL \\
\midrule
Swin & 73.93$\pm$2.2$^*$ & 66.97$\pm$3.9 & 68.12$\pm$1.7 & 66.76$\pm$4.0 & 66.76$\pm$4.0 & 85.89$\pm$3.2 & 84.11$\pm$2.8$^*$ & 0.81$\pm$0.8 \\
ResNet & 71.37$\pm$4.4$^{**}$ & 64.19$\pm$2.2$^{**}$ & 65.16$\pm$3.0 & 63.56$\pm$1.3 & 63.56$\pm$1.3 & 84.33$\pm$4.4 & 82.05$\pm$0.9$^{**}$ & 0.82$\pm$0.7 \\
DenseNet & 74.64$\pm$1.1 & 68.62$\pm$3.7 & 69.08$\pm$2.6 & 68.58$\pm$3.8 & 68.58$\pm$3.8 & 86.66$\pm$1.1 & 86.81$\pm$4.2 & 1.01$\pm$1.0 \\
EfficientNetV2 & 73.65$\pm$3.2 & 68.38$\pm$0.8 & 69.49$\pm$4.5 & 69.06$\pm$2.8 & 69.06$\pm$2.8 & 86.42$\pm$2.1 & 88.06$\pm$3.3 & 1.60$\pm$1.4 \\
ViT & 64.67$\pm$1.8$^{***}$ & 55.12$\pm$4.8$^{***}$ & 59.39$\pm$2.5 & 55.25$\pm$1.8 & 55.25$\pm$1.8 & 79.74$\pm$3.8 & 78.13$\pm$4.2$^{***}$ & 0.81$\pm$0.9 \\
DeiT & 73.36$\pm$4.2 & 67.71$\pm$1.8 & 67.42$\pm$3.9 & 68.14$\pm$0.8 & 68.14$\pm$0.8 & 86.46$\pm$1.1 & 87.17$\pm$2.5 & 1.30$\pm$1.2 \\
CLIP & 66.10$\pm$3.7$^{**}$ & 59.75$\pm$2.5 & 60.78$\pm$4.3 & 60.24$\pm$1.6 & 60.24$\pm$1.6 & 81.61$\pm$0.7 & 76.81$\pm$3.4$^{**}$ & 1.05$\pm$1.0 \\
Ours & \textbf{78.06}$\pm$2.4 & \textbf{72.63}$\pm$1.3 & \textbf{73.55}$\pm$3.7 & \textbf{72.57}$\pm$0.9 & 72.57$\pm$0.9 & \textbf{88.43}$\pm$4.1 & \textbf{89.77}$\pm$2.8 & 1.57$\pm$3.5 \\
\bottomrule
\end{tabular}
\\
\vspace{2pt}
\scriptsize\raggedright
\end{minipage}
\end{table*}

\subsubsection{Additional Class-wise and Binary Experiments}

To further examine the classification performance across different pathological subtypes, we conducted class-wise comparison experiments on a class-balanced subset. The class-wise results for FA, benign PT, and borderline PT are shown in Table~\ref{tab:subcategory_acc}.

On this balanced subset, the full model proposed in this paper achieved an overall ACC of 74.17\%, the highest overall result among the compared methods. The class-level results show that our method achieves a high recognition accuracy for FA, whilst the results for benign PT and borderline PT exhibit more pronounced class-specific differences.

In addition to the three-class classification task, we further conducted supplementary FA--PT binary classification experiments, merging benign PT and borderline PT into a single PT class, to evaluate the model's overall ability to distinguish between the two types of fibroepithelial lesions: FA and PT. The results of the binary classification are shown in Table~\ref{tab:binary_fa_pt}.

Our method achieved an overall ACC of 89.60\% in the FA--PT binary classification task, outperforming the compared image-based models and different modality combination configurations. This supplementary experiment further evaluated the model's overall discriminative capability for FA and PT from a binary classification perspective.

\subsubsection{Generalisation Evaluation on the PetFinder Dataset}

To further assess whether the proposed framework is limited to breast ultrasound scenarios, we conducted additional experiments on the publicly available PetFinder dataset\cite{KagglePetFinder2019}, which has also been included in the multimodal benchmark for evaluating AutoMM\cite{Tang2024AutoMM}. The PetFinder dataset originates from the PetFinder.my adoption prediction competition and contains heterogeneous information such as pet images, textual descriptions and structured attributes; it is therefore well-suited for evaluating multimodal learning methods.

PetFinder \cite{KagglePetFinder2019} differs significantly from FAPT-M in both application domain and data characteristics. Consequently, this experiment is not conducted as part of the main breast ultrasound comparison, but rather as an additional cross-dataset evaluation aimed at verifying whether the proposed multimodal information flow strategy can adapt to distinctly different multimodal classification tasks.

As the only metric reported by AutoMM on this dataset is the weighted F1-score, this paper adopts the same metric for evaluation to ensure direct comparability of the results. AutoMM achieved a result of 40.8 \%, whilst the method proposed in this paper attained 43.73\%, representing an absolute improvement of 2.93\%. This experiment serves as supplementary validation across datasets, providing additional empirical evidence for the applicability of the proposed framework in non-medical multimodal tasks.

\subsection{Ablation Studies}

To evaluate the contribution of different design elements within the proposed framework to the final performance, this paper conducts both module-level ablation experiments and modal contribution experiments.

\subsubsection{Module-Level Ablation}

The module-level ablation experiments primarily evaluate key designs within the ‘condition–interact–retain’ information flow, including the retention of both original and context dual paths, the bottleneck layer, the CLS token, the gating mechanism, and clinical conditional modulation. The clinical conditioning component is based on the feature-level affine modulation approach \cite{Perez2018FiLM}, whilst cross-modal interaction is realised via the Transformer’s self-attention mechanism \cite{Vaswani2017Transformer}.

The complete ablation results are shown in Table~\ref{tab:module_ablation_full}. In the ‘Only-Fused’ configuration, only the representations updated via cross-modal interaction are retained, whilst the original modal representations are no longer utilised in conjunction. Both its ACC and F1-score are lower than those of the full model, providing experimental support for the strategy of retaining a combined representation of original modal evidence and contextual information.

Removing the bottleneck also results in a decline in the model’s overall performance. Removing the CLS token leads to a slight improvement in the F1-score and recall, but a decrease in ACC and AUC, indicating that this component involves a trade-off between different metrics.

When the Gate or FiLM conditional modulation was removed, the model’s ACC and F1-score were both lower than those of the full model. The full model achieved the highest ACC and AUC in this set of experiments. Overall, Table~\ref{tab:module_ablation_full} demonstrates that each module contributes to the full model’s overall classification performance to varying degrees, but no single component guarantees consistent improvements across all metrics.

\begin{table*}[!t]
\centering
\caption{Module-Level Ablation Experiment Results on the FAPT-M Dataset}
\fontsize{8}{10}\selectfont
\setlength{\tabcolsep}{4pt}
\begin{minipage}{\linewidth}
\centering
\begin{tabular}{l c c c c c c c c}
\toprule
Method & ACC(\%) & F1-Score(\%) & PRE(\%) & REC(\%) & Sensitivity(\%) & Specificity(\%) & AUC & NLL \\
\midrule
Only-Fused & 73.79$\pm$1.2 & 69.27$\pm$3.8 & 68.66$\pm$2.5 & 70.40$\pm$4.1 & 70.40$\pm$4.1 & 87.37$\pm$0.9 & 88.27$\pm$3.2 & 1.36$\pm$2.8 \\
w/o Bottleneck & 75.50$\pm$2.7 & 70.69$\pm$1.5 & 70.60$\pm$4.3 & 70.84$\pm$0.8 & 70.84$\pm$0.8 & 87.17$\pm$3.5 & 89.00$\pm$2.1 & 1.19$\pm$1.4 \\
w/o CLS Token & 76.64$\pm$4.0 & \textbf{72.90}$\pm$2.9 & 72.20$\pm$1.7 & \textbf{74.09}$\pm$3.3 & \textbf{74.09}$\pm$3.3 & 88.69$\pm$2.2 & 89.32$\pm$0.7 & 1.29$\pm$4.2 \\
w/o Gate & 75.78$\pm$3.1 & 70.70$\pm$4.5 & 70.47$\pm$2.3 & 71.13$\pm$1.9 & 71.13$\pm$1.9 & 87.75$\pm$0.6 & 89.38$\pm$3.8 & 1.56$\pm$2.5 \\
w/o FiLM& 75.21$\pm$0.8 & 71.47$\pm$3.2 & 70.79$\pm$4.0 & 72.93$\pm$2.6 & 72.93$\pm$2.6 & 87.97$\pm$1.8 & 89.76$\pm$3.0 & 1.30$\pm$1.1 \\
Ours & \textbf{78.06}$\pm$2.4 & 72.63$\pm$1.3 & \textbf{73.55}$\pm$3.7 & 72.57$\pm$0.9 & 72.57$\pm$0.9 & \textbf{88.43}$\pm$4.1 & \textbf{89.77}$\pm$2.8 & 1.57$\pm$3.5 \\
\bottomrule
\end{tabular}
\\
\vspace{2pt}
\scriptsize\raggedright
\end{minipage}
\label{tab:module_ablation_full}
\end{table*}

\subsubsection{Modality Contribution}

To analyse the contributions of the three modalities—images, ultrasound descriptions and structured clinical information—this paper further compares four different input configurations: Image Only, Image+Clinical, Image+Text, and the full model utilising all three modalities simultaneously.

All configurations employ the same base network architecture and training strategy, with only the combination of input modalities altered. The complete experimental results are shown in Table~\ref{tab:modality_ablation_full}.

Compared with the model using images alone, the inclusion of structured clinical information or ultrasound descriptions resulted in improvements in both ACC and F1-score. The ACC for Image+Clinical increased from 73.65\% to 74.50\%, whilst Image+Text achieved a further improvement to 76.78\%.

The full tri-modal model achieved an ACC of 78.06\% and an F1-score of 72.63\%, attaining the highest ACC, F1-score, PRE, REC and specificity among the compared modal configurations. Although ‘Image+Text’ achieved a higher AUC and ‘Image+Clinical’ achieved a lower NLL, the overall results in Table~\ref{tab:modality_ablation_full} indicate that the combination of all three modalities delivers the most stable overall performance across key classification metrics.

\begin{table*}[!t]
\centering
\caption{Modality Contribution Ablation Experiment Results on the FAPT-M DataSset}
\fontsize{8}{10}\selectfont
\setlength{\tabcolsep}{4pt}
\begin{minipage}{\linewidth}
\centering
\begin{tabular}{l c c c c c c c c}
\toprule
Method & ACC(\%) & F1-Score(\%) & PRE(\%) & REC(\%) & Sensitivity(\%) & Specificity(\%) & AUC(\%) & NLL \\
\midrule
Image Only & 73.65$\pm$4.5 & 68.39$\pm$1.9 & 67.94$\pm$3.1 & 69.18$\pm$2.8 & 69.18$\pm$2.8 & 87.35$\pm$0.8 & 86.94$\pm$4.2 & 1.39$\pm$1.2 \\
Image+Clinical & 74.50$\pm$3.1 & 69.51$\pm$4.0 & 69.01$\pm$2.0 & 70.25$\pm$1.6 & 70.25$\pm$1.6 & 87.55$\pm$3.0 & 88.82$\pm$1.3 & 0.94$\pm$0.9 \\
Image+Text & 76.78$\pm$0.8 & 72.56$\pm$2.4 & 72.35$\pm$3.5 & 73.16$\pm$4.3 & 73.16$\pm$4.3 & 88.10$\pm$1.9 & 90.31$\pm$2.1 & 1.14$\pm$1.0 \\
Ours & \textbf{78.06}$\pm$2.4 & \textbf{72.63}$\pm$1.3 & \textbf{73.55}$\pm$3.7 & 72.57$\pm$0.9 & 72.57$\pm$0.9 & \textbf{88.43}$\pm$4.1 & \textbf{89.77}$\pm$2.8 & 1.57$\pm$3.5 \\
\bottomrule
\end{tabular}
\\
\vspace{2pt}
\scriptsize\raggedright
\end{minipage}
\label{tab:modality_ablation_full}
\end{table*}

\subsection{Class-Balanced Analysis}

Because FAPT-M is derived from real-world clinical data, the number of cases varies across pathological categories. To further assess whether the class distribution affects model performance, this study constructed a class-balanced subset comprising 200 ultrasound images each for FA, benign PT and borderline PT.

Sample extraction was carried out within predefined, mutually exclusive patient-level data partitions, without altering the assignment of any cases to the training, validation or test partitions. All comparison configurations were retrained and evaluated on the same balanced subset.

First, the overall performance of different modal input combinations on the balanced subset was compared, with the results shown in Table~\ref{tab:modality_ablation_general}. Under balanced data conditions, the full tri-modal model achieved a further improvement to 74.17\%. The full model also attained the highest results across the primary metrics.

Subsequently, module-level ablation experiments were repeated on the same balanced subset, with the results shown in Table~\ref{tab:module_ablation_general_200}. These results are consistent with the overall trends observed in the previous module-level ablation, providing further support for the effectiveness of the full information flow design after controlling for class distribution.

Finally, to assess the impact of different configurations on pathological subtypes under class-balanced conditions, we also report class-level accuracy. As shown in Table~\ref{tab:modality_ablation_class_acc}, different modalities and module configurations exhibit varying class preferences for FA, benign PT and borderline PT. The complete model achieved an ACC of 92.50\% for FA; the Image+Text configuration obtained the highest result for borderline PT; whilst benign PT remained the relatively more challenging class.

In summary, after balancing the number of samples across classes, the full model in this study still maintains the highest overall ACC, indicating that the performance advantages observed in the main experiment are not solely attributable to the original class distribution.

\begin{table*}[!t]
\centering
\caption{Comparison with independent multimodal baselines on the FAPT-M dataset}
\label{tab:multimodal_baselines}
\fontsize{8}{10}\selectfont
\setlength{\tabcolsep}{3.2pt}

\begin{minipage}{\linewidth}
\centering

\begin{tabular}{l c c c c c c c c}
\toprule
Method &
ACC (\%) &
F1-Score (\%) &
PRE (\%) &
REC (\%) &
Sensitivity (\%) &
Specificity (\%) &
AUC (\%) &
NLL \\
\midrule

MMBCD &
66.76$\pm$1.24 &
59.85$\pm$2.31 &
60.29$\pm$3.87 &
59.65$\pm$1.02 &
59.65$\pm$1.02 &
83.03$\pm$0.98 &
80.14$\pm$2.76 &
1.9027 \\

LMF &
70.82$\pm$2.54 &
62.60$\pm$1.78 &
68.53$\pm$3.45 &
60.77$\pm$4.62 &
60.77$\pm$4.62 &
82.19$\pm$2.09 &
81.18$\pm$1.56 &
0.7984 \\

AutoMM-LF &
61.13$\pm$3.21 &
54.23$\pm$1.44 &
54.86$\pm$2.87 &
54.50$\pm$0.79 &
54.50$\pm$0.79 &
79.62$\pm$1.91 &
73.92$\pm$2.40 &
2.1086 \\

Ours & \textbf{78.06}$\pm$2.4 & \textbf{72.63}$\pm$1.3 & \textbf{73.55}$\pm$3.7 & \textbf{72.57}$\pm$0.9 & \textbf{72.57}$\pm$0.9 & \textbf{88.43}$\pm$4.1 & \textbf{89.77}$\pm$2.8 & 1.57$\pm$3.5 \\


\bottomrule
\end{tabular}

\vspace{3pt}

\end{minipage}
\end{table*}

\begin{table*}[!t]
\centering
\caption{Class-wise Comparison and Ablation Experiment Results on the Balanced FAPT-M Subset}
\fontsize{8}{10}\selectfont
\setlength{\tabcolsep}{3pt}
\begin{minipage}{\linewidth}
\centering
\begin{tabular}{l c c c c c c c c c}
\toprule
Category & ResNet(\%) & DenseNet(\%) & EfficientNetV2(\%) & Swin(\%) & DeiT(\%) & Image(\%) & Image+Clinical(\%) & Image+Text(\%) & Ours(\%) \\
\midrule
FA & 75.00$\pm$4.5 & 87.50$\pm$2.2 & 72.50$\pm$1.7 & 85.00$\pm$3.5 & 80.00$\pm$4.0 & 85.00$\pm$2.5 & 87.50$\pm$1.2 & 85.00$\pm$3.5 & \textbf{92.50}$\pm$0.8 \\
B & 42.50$\pm$3.8 & 47.50$\pm$4.5 & 52.50$\pm$2.2 & 65.00$\pm$1.3 & 40.00$\pm$3.9 & 55.00$\pm$3.0 & 67.50$\pm$4.4 & 50.00$\pm$2.8 & \textbf{60.00}$\pm$1.7 \\
BD & 72.50$\pm$2.7 & 70.00$\pm$1.8 & 45.00$\pm$3.8 & 52.50$\pm$4.2 & 80.00$\pm$0.4 & 60.00$\pm$2.7 & 55.00$\pm$3.0 & \textbf{80.00}$\pm$1.2 & 70.00$\pm$2.9 \\
Total & 63.33$\pm$1.9 & 68.33$\pm$2.4 & 56.67$\pm$3.9 & 67.50$\pm$0.5 & 66.67$\pm$1.8 & 66.67$\pm$2.8 & 70.00$\pm$3.0 & 71.67$\pm$4.1 & \textbf{74.17}$\pm$3.8 \\
\bottomrule
\end{tabular}
\\
\vspace{2pt}
\scriptsize\raggedright
\end{minipage}
\label{tab:subcategory_acc}
\end{table*}

\begin{table*}[!t]
\centering
\caption{Binary FA--PT Classification  Comparison and Ablation Experiment Results on the FAPT-M Dataset}
\fontsize{8}{10}\selectfont
\setlength{\tabcolsep}{3pt}
\begin{minipage}{\linewidth}
\centering
\begin{tabular}{l c c c c c c c c c}
\toprule
Category & ResNet(\%) & DenseNet(\%) & EfficientNetV2(\%) & Swin(\%) & DeiT(\%) & Image(\%) & Image+Clinical(\%) & Image+Text(\%) & Ours(\%) \\
\midrule
FA & 82.15$\pm$3.1 & 85.30$\pm$4.7 & 85.36$\pm$2.7 & 85.04$\pm$1.8 & 83.73$\pm$3.0 & 84.51$\pm$4.9 & 81.99$\pm$2.1 & \textbf{89.76}$\pm$1.2 & 88.45$\pm$2.4 \\
PT & 82.87$\pm$2.0 & 87.54$\pm$1.5 & 82.15$\pm$3.1 & 87.54$\pm$2.5 & 87.23$\pm$0.6 & 89.35$\pm$1.3 & \textbf{93.46}$\pm$3.8 & 86.92$\pm$2.7 & \textbf{90.97}$\pm$1.1 \\
Total & 82.48$\pm$4.0 & 86.32$\pm$2.6 & 83.62$\pm$1.9 & 86.18$\pm$3.6 & 85.33$\pm$2.8 & 86.75$\pm$1.5 & 87.18$\pm$4.5 & 88.46$\pm$3.4 & \textbf{89.60}$\pm$2.2 \\
\bottomrule
\end{tabular}
\\
\vspace{2pt}
\scriptsize\raggedright
\end{minipage}
\label{tab:binary_fa_pt}
\end{table*}

\begin{table*}[!t]
\centering
\caption{Modality Contribution Ablation Study on Balanced Subset}
\fontsize{8}{10}\selectfont
\setlength{\tabcolsep}{4pt}
\begin{minipage}{\linewidth}
\centering 
\begin{tabular}{l c c c c c c c c}
\toprule
Method & ACC(\%) & F1-Score(\%) & PRE(\%) & REC(\%) & Sensitivity(\%) & Specificity(\%) & AUC & NLL \\
\midrule
Image Only & 66.67$\pm$2.1 & 66.41$\pm$3.8 & 66.53$\pm$1.9 & 66.67$\pm$2.1 & 66.67$\pm$2.1 & 83.33$\pm$4.2 & 82.02$\pm$3.5 & 1.08$\pm$0.9 \\
Image+Clinical & 70.00$\pm$1.5 & 69.58$\pm$2.9 & 69.43$\pm$4.1 & 70.00$\pm$1.5 & 70.00$\pm$1.5 & 85.00$\pm$2.5 & 84.71$\pm$0.8 & 0.69$\pm$1.2 \\
Image+Text & 71.67$\pm$4.3 & 71.30$\pm$1.7 & 72.56$\pm$2.5 & 71.67$\pm$4.3 & 71.67$\pm$4.3 & 85.83$\pm$3.4 & 82.10$\pm$4.0 & 1.01$\pm$0.7 \\
Ours & \textbf{74.17}$\pm$0.9 & \textbf{73.90}$\pm$4.5 & \textbf{74.42}$\pm$3.2 & \textbf{74.17}$\pm$0.9 & \textbf{74.17}$\pm$0.9 & \textbf{87.08}$\pm$1.3 & \textbf{87.31}$\pm$2.9 & 1.67$\pm$1.5 \\
\bottomrule
\end{tabular}
\\ 
\vspace{2pt} 
\scriptsize\raggedright 
\end{minipage}
\label{tab:modality_ablation_general}
\end{table*}

\begin{table*}[!t]
\centering
\caption{Module-Level Ablation Study on Balanced Subset}
\fontsize{7.5}{9}\selectfont
\setlength{\tabcolsep}{5pt}
\begin{minipage}{\linewidth}
\centering
\begin{tabular}{l c c c c c c c c}
\toprule
Method & ACC(\%) & F1-Score(\%) & PRE(\%) & REC(\%) & Sensitivity(\%) & Specificity(\%) & AUC & NLL \\
\midrule
Only-Fused & 70.83$\pm$3.6 & 70.34$\pm$1.2 & 70.11$\pm$4.4 & 70.83$\pm$3.6 & 70.83$\pm$3.6 & 85.42$\pm$0.5 & 86.85$\pm$2.0 & 1.37$\pm$1.3 \\
w/o Bottleneck & 70.83$\pm$2.6 & 71.47$\pm$4.0 & 73.04$\pm$1.5 & 70.83$\pm$2.6 & 70.83$\pm$2.6 & 85.42$\pm$3.5 & 83.78$\pm$0.9 & 0.75$\pm$0.8 \\
w/o CLS Token & 71.67$\pm$1.3 & 71.94$\pm$3.8 & 72.71$\pm$2.6 & 71.67$\pm$1.3 & 71.67$\pm$1.3 & 85.83$\pm$4.3 & 85.96$\pm$3.1 & 1.27$\pm$1.2 \\
w/o Gate & 72.50$\pm$4.2 & 72.21$\pm$0.7 & 72.18$\pm$3.7 & 72.50$\pm$4.2 & 72.50$\pm$4.2 & 86.25$\pm$2.3 & 86.07$\pm$1.0 & 0.69$\pm$1.4 \\
w/o FiLM & 71.67$\pm$0.8 & 71.33$\pm$2.8 & 71.12$\pm$1.1 & 71.67$\pm$0.8 & 71.67$\pm$0.8 & 85.83$\pm$3.4 & 84.60$\pm$2.3 & 0.73$\pm$0.9 \\
Ours & \textbf{74.17}$\pm$0.9 & \textbf{73.90}$\pm$4.5 & \textbf{74.42}$\pm$3.2 & \textbf{74.17}$\pm$0.9 & \textbf{74.17}$\pm$0.9 & \textbf{87.08}$\pm$1.3 & \textbf{87.31}$\pm$2.9 & 1.67$\pm$1.5 \\
\bottomrule
\end{tabular}
\\
\vspace{2pt}
\scriptsize\raggedright
\end{minipage}
\label{tab:module_ablation_general_200}
\end{table*}


\begin{table}[!t]
\centering
\caption{Ablation Study on Balanced Subset (Class-wise ACC, 200 Samples per Subcategory)}
\label{tab:modality_ablation_class_acc}
\renewcommand{\arraystretch}{1.1}
\makebox[\linewidth][c]{%
\resizebox{0.45\linewidth}{!}{%
\begin{tabular}{l c c c}
\toprule
Method & FA(\%) & B(\%) & BD(\%) \\
\midrule
Image Only & 85.00$\pm$4.5 & 55.00$\pm$3.2 & 60.00$\pm$1.7 \\
Image+Clinical & 87.50$\pm$3.9 & 67.50$\pm$4.4 & 55.00$\pm$2.8 \\
Image+Text & 85.00$\pm$2.3 & 50.00$\pm$1.9 & 80.00$\pm$3.1 \\
Only-Fused & 92.50$\pm$4.5 & 67.50$\pm$2.7 & 52.50$\pm$3.0 \\
w/o Bottleneck & 87.50$\pm$1.2 & 62.50$\pm$4.8 & 72.50$\pm$2.4 \\
w/o CLS Token & 87.50$\pm$2.9 & 60.00$\pm$1.8 & 67.50$\pm$0.7 \\
w/o Gate & 90.00$\pm$3.8 & 65.00$\pm$2.8 & 62.50$\pm$4.1 \\
w/o FiLM & 87.50$\pm$4.2 & 72.50$\pm$1.4 & 55.00$\pm$3.9 \\
Ours & 92.50$\pm$4.0 & 60.00$\pm$2.6 & 70.00$\pm$3.8 \\
\bottomrule
\end{tabular}
}%
}
\end{table}

\section{Discussion}

\subsection{Comparison With Baselines and Design Implications}

This study focuses on the fine-grained classification task of fibroepithelial lesions, which present a high degree of clinical differentiation difficulty. Comparison results with single-modality image classification models are shown in Table~\ref{tab:image_baselines}, whilst those with independent multimodal models are presented in Table~\ref{tab:multimodal_baselines}. In particular, MMBCD, LMF and AutoMM-LF represent multimodal modelling of breast imaging, low-rank fusion and generalised multimodal learning approaches, respectively; however, their performance on FAPT-M is inferior to that of the multimodal model proposed in this study. Therefore, the overall results indicate that model performance depends not only on whether additional modalities are introduced, but is also closely related to the way in which information across different modalities is organised.

Based on this consideration, the methodological innovations in this paper primarily focus on the flow of multimodal information. The proposed ‘condition–interact–retain’ mechanism first utilises structured clinical information to conditionally adjust image and text representations, enabling patient-specific clinical context to influence feature representations before cross-modal interaction; subsequently, different modalities interact via a unified token representation; and finally, both the original modal information and the context information updated through interaction are retained simultaneously.

Module-level ablation in Table~\ref{tab:module_ablation_full} further supports this design. Retaining only the post-interaction representations weakens overall classification performance, indicating that relying solely on fused representations may result in the loss of some modality-specific information, whilst jointly retaining the original and contextual representations helps to preserve discriminative information. Ablation results for FiLM and Gate similarly indicate that clinical conditional modulation has a positive effect on the complete framework. At the same time, the impact of different components on various evaluation metrics is not entirely consistent; for example, removing certain components may improve one metric whilst simultaneously reducing others. Therefore, the advantages of the complete framework are more likely to stem from the synergy between multiple design elements, rather than any single module generating independent gains under all conditions.

\subsection{Modality Contributions and Subtype Performance}

The results of different modality combinations are shown in Table~\ref{tab:modality_ablation_full}. The ablation results indicate that both text and structured clinical information complement image representations, supporting the use of clinical variables as conditional signals for image and text features.

Class-wise results in Table~\ref{tab:subcategory_acc} further show that modality contributions vary across pathological subtypes. FA is identified relatively consistently, whereas benign and borderline PT remain more challenging. The auxiliary FA--PT binary results in Table~\ref{tab:binary_fa_pt} further suggest that the main difficulty lies not only in distinguishing FA from PT, but also in differentiating benign from borderline PT, highlighting the value of the fine-grained task beyond conventional FA--PT binary classification.

Class-balanced analysis further controls for the influence of class distribution. Results for modality contribution, module-level ablation, and class-wise accuracy are reported in Tables~\ref{tab:modality_ablation_general}, \ref{tab:module_ablation_general_200}, and \ref{tab:modality_ablation_class_acc}, respectively. The complete model remains competitive under the balanced setting, suggesting that its advantage cannot be explained solely by class imbalance.

Finally, we adapted the framework to the PetFinder multimodal dataset and compared it with AutoMM. Despite substantial differences between PetFinder and FAPT-M, the adapted framework remained competitive, providing preliminary evidence of cross-dataset generalizability.

\subsection{Interpretability and Limitations}

\begin{figure}[!t]
\centering
\includegraphics[width=0.6\columnwidth]{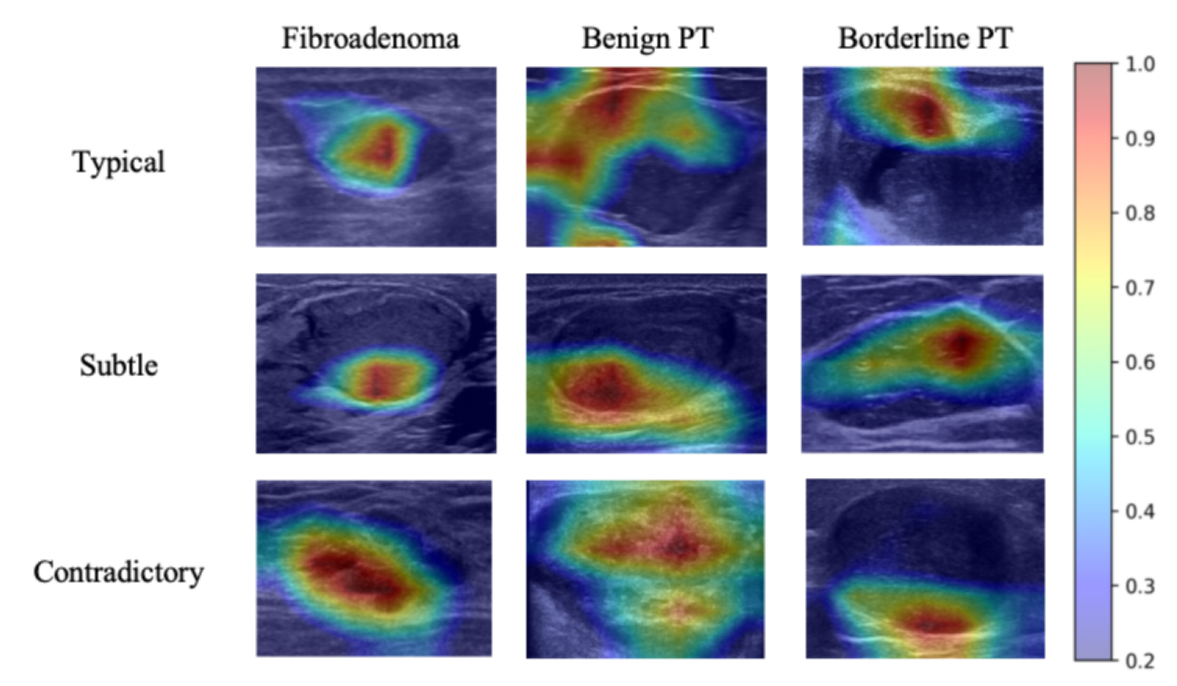}
\caption{Representative Grad-CAM visualizations of FA and PT cases. Red regions indicate stronger contributions to predictions, while blue regions indicate weaker contributions.}
\label{fig3}
\end{figure}

As shown in Fig.~\ref{fig3}, the Grad-CAM \cite{Selvaraju201} visualisation results indicate that the primary response regions are concentrated within the lesion and near its boundaries; in cases where image differences are subtle, or the presentation is atypical, the model’s regions of interest generally remain within lesion-related areas. This suggests that some of the visual discriminative information learnt by the model exhibits a certain degree of spatial consistency with clinically relevant regions.

This study still has several limitations. Firstly, FAPT-M is based on single-centre data, which is prone to centre-specific distributions; therefore, validation on multi-centre data is recommended prior to clinical application. Furthermore, in future work, we intend to incorporate additional modalities and evaluate the framework on a wider range of publicly available multi-modal datasets to investigate the decision-making contributions of different modalities further.

\section{Conclusion}

We constructed the FAPT-M multimodal dataset for the fine-grained differentiation of FA, benign PT, and borderline PT, and proposed a clinically conditioned multimodal learning framework based on a condition--interact--retain information flow. The proposed method uses structured clinical information to condition image and text representations, integrates multimodal information through cross-modal interaction, and retains both original and contextualized representations for final classification. Experimental results demonstrate that the proposed method consistently outperforms both image-based approaches and commonly used multimodal methods, while ablation studies further validate the effectiveness of its key design components. Experiments on a general-purpose dataset further demonstrate the generalizability of the proposed method. Future multi-center and prospective studies are still required to further assess its clinical applicability.

\end{document}